\pdfoutput=1

\documentclass[11pt]{article}

\usepackage{acl}

\usepackage{times}
\usepackage{latexsym}

\usepackage[T1]{fontenc}

\usepackage[utf8]{inputenc}

\usepackage{microtype}

\usepackage{smile}

\newcommand{\ours}{{{ETC}}} 

\title{Short Text Pre-training with Extended Token Classification for E-commerce Query Understanding}

\author{\makecell{Haoming Jiang$^*$, Tianyu Cao, Zheng Li, Chen Luo, Xianfeng Tang \\
  Qingyu Yin, Danqing Zhang, Rahul Goutam, Bing Yin} \\
  Amazon Search \\
  \texttt{jhaoming@amazon.com} \\ 
  }
  
\begin{document}
\maketitle

\newcommand{\zheng}[1]{\textcolor{teal}{\textbf{[ZHENG: #1]}}}
\newcommand{\yqy}[1]{\textcolor{magenta}{\textbf{[Qingyu: #1]}}}
\newcommand{\caoty}[1]{\textcolor{green}{\textbf{[Qingyu: #1]}}}
\newcommand{\chen}[1]{\textcolor{blue}{\textbf{[Chen: #1]}}}
\newcommand{\dq}[1]{\textcolor{red}{\textbf{[Danqing: #1]}}}

\begin{abstract}

E-commerce query understanding is the process of inferring the shopping intent of customers by extracting semantic meaning from their search queries. 
The recent progress of pre-trained masked language models (MLM) in natural language processing is extremely attractive for developing effective query understanding models. 
Specifically, MLM learns contextual text embedding via recovering the masked tokens in the sentences. 
Such a pre-training process relies on the sufficient contextual information. It is, however, less effective for search queries, which are usually short text. 
When applying masking to short search queries, most contextual information is lost and the intent of the search queries may be changed. 
To mitigate the above issues for MLM pre-training on search queries, we propose a novel pre-training task specifically designed for short text, called Extended Token Classification (\ours). 
Instead of masking the input text, our approach extends the input by inserting tokens via a generator network, and trains a discriminator to identify which tokens are inserted in the extended input.
We conduct experiments in an E-commerce store to demonstrate the effectiveness of {\ours}. 
\end{abstract}
\section{Introduction}

Query Understanding (QU) plays an essential role in E-commerce shopping platform, where it extracts the shopping intents of the customers from their search queries.
Traditional approaches usually rely on handcrafted features or rules \cite{henstock2001toward}, which only have limited coverage. 
More recently, deep learning models are proposed to improve the the generalization performance of QU models \citep{nigam2019semantic,lin2020light}. 
These methods usually train a deep learning model from scratch, which requires a large amount of manually labeled data. Annotating a large number of queries can be expensive, time-consuming, and prone to human errors. Therefore, the labeled data is often limited.

To achieve better model performance with limited data, researchers resorted to the masked language model (MLM) pre-training with large amount of unlabeled open-domain data  \citep{devlin2018bert,liu2019roberta,jiang2019smart, He2021DeBERTaV3ID} and achieved the state-of-the-art performance in QU tasks \citep{kumar2019shareable, jiang2021named, zhang2021queaco, li2021metats}.
However, open-domain pre-trained models  can only provide limited semantic and syntax information for QU tasks in E-commerce search domain.
In order to capture domain-specific information, \citet{lee2020biobert,gururangan2020don,gu2021domain} propose to pre-train the MLM on a large in-domain unlabeled data either initialized randomly or from a public pre-trained checkpoint. 

\begin{figure}[htb]
    \centering
    \includegraphics[width=0.5\textwidth]{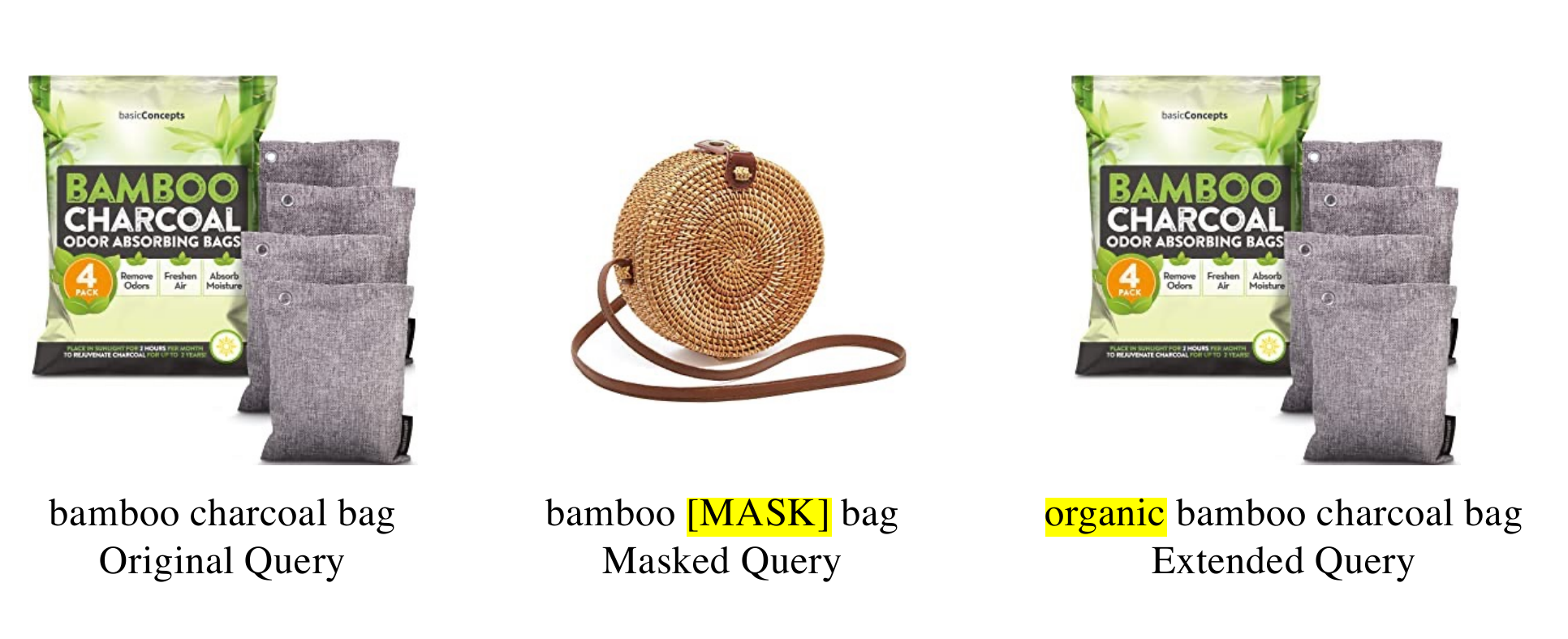}
    \caption{Original Query vs. Masked Query vs. Extended Query. `bamboo charcoal bag' is a bag of `bamboo charcoal', while `bamboo bag' is a bag made of bamboo.  Masking out `charcoal' will completely change the user's search intent. On the contrary, extending the query to `organic bamboo charcoal bag' does not change the user's search intent even though the combination of `organic' and `bamboo charcoal bag' is not common. }
    \label{fig:vs_exp}
    \vspace{-0.1in}
\end{figure}

Although the search query domain specific MLM can adapt to the search query distribution in some extent, it is not effective in capturing the contextual information of the search queries due to the short length of the queries. There are two major challenges: 

\noindent~$\bullet$ 
MLM \cite{devlin2018bert} randomly replace tokens in the text by \texttt{[MASK]} tokens and train the model to recover them with a low masking probability (e.g., $15\%$ in \citet{devlin2018bert,liu2019roberta}). 
Since the length of search queries is short, there will be many queries having no masked tokens during training. Even though we can ensure each query to be masked for at least one token, the percentage of mask tokens will be way much higher and thus loss much context information; 

\noindent~$\bullet$ Masking out tokens may significantly change the intent of the search queries. Figure~\ref{fig:vs_exp} shows an example of masked token changing the intent of the query.

\begin{figure*}[!htb]
    \centering
    \includegraphics[width=0.95\textwidth]{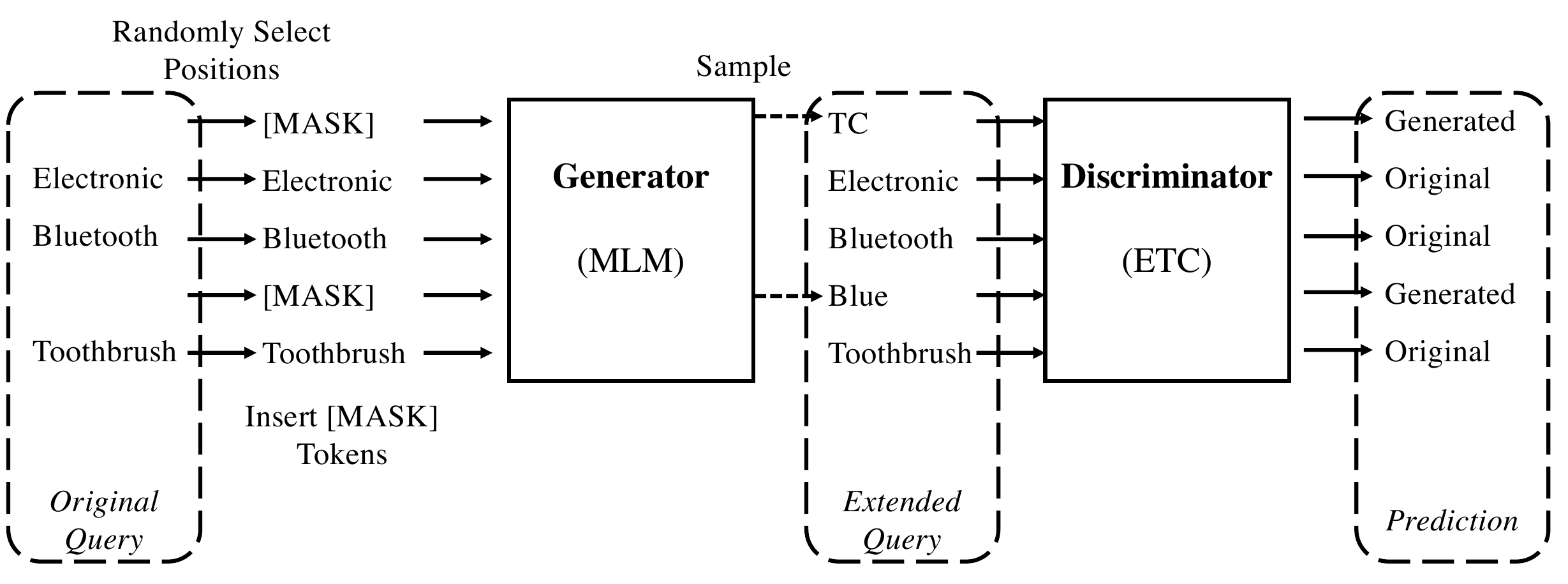}
    \caption{An overview of extended token classification. 
    In this example, the discriminator should be able to identify the inserted tokens by telling that ``TC Electronic'' is a brand but not for toothbrush and ``Bluetooth'' is much more common than ``Blue Toothbrush''.
    After pre-training, we keep the discriminator as the encoder for query representation. 
    }
    \vspace{-0.2in}
    \label{fig:framework}
\end{figure*}

In this paper, we propose a new pre-training task --- \textbf{E}xtended \textbf{T}oken \textbf{C}lassification (\textbf{\ours}) to mitigate the above-mentioned issues for search query pre-training. 
Instead of masking out tokens in the search query and training the model to recover the tokens, we extend the search query and train the model to identify which tokens are extended. 
The extended query is generated by inserting tokens with a generator, which is a pre-trained masked language model. The generator takes query with randomly inserted \texttt{[MASK]} tokens as the input and fill in the blanks by its prediction. 
There are several benefits of {\ours}: 

\noindent~$\bullet$ It turns the language modeling task into a binary classification task on all tokens, which makes the model easier to train;\\
\noindent~$\bullet$ All samples will be used to train the model, even when the probability of inserting tokens is low;\\
\noindent~$\bullet$ Since the generator has already been pre-trained, the extended queries alter the meaning of the search query less frequently. 

We conduct experiments on an E-commerce store to demonstrate the effectiveness of {\ours}. We conduct fine-tuning experiments in a wide range of query understanding tasks including three classification tasks, one sequence labeling task, and one text generation task. We show that {\ours} outperforms open-domain pre-trained models, and search query domain specific pre-trained MLM model and ELECTRA \citep{clark2020electra} model.
\section{Background}

Masked Language Modeling ({MLM}) pre-training is first introduced in \citet{devlin2018bert} to learn contextual word representations with a large transformer model \citep{vaswani2017attention}. 
Given a sequence of tokens $\bx = [x_1, ..., x_n]$, \citet{devlin2018bert} corrupt it into $\bx^{\rm mask}$ by masking $15\%$ of its tokens at random:
\begin{align*}
    &m_i \sim {\rm Binomial}(0.15), ~ {\rm for} ~ i \in [0,...,n] \\
    \bx^{\rm mask} &= {\rm REPLACE}(\bx, [m_1,...,m_n], \texttt{[MASK]})
\end{align*}

\citet{devlin2018bert} then train a transformer-based language model $G$ parameterized by $\theta$ to reconstruct $\bx$ conditioned on $\bx^{\rm mask}$:
\begin{align*}
   \min_{\theta} \EE  \sum_{t = 1}^n \ind(m_t=1) p_{G} ( x_t | \bx^{\rm mask} ), 
\end{align*}
where $p_{G} ( x_t | \bx^{\rm mask} )$ denotes the predicted probability of the $t$-th token being $x_t$ given $\bx^{\rm mask}$. 

\citet{devlin2018bert} also introduced a next sentence prediction (NSP) pre-training task, which is shown to be not very effective in a later work \citep{liu2019roberta}. In this paper, we do not discuss the NSP pre-training task.

\section{Method}

{\ours} adopts two transformer-based neural networks: a generator $G$ and a discriminator $D$. A raw text input is first inserted with some \texttt{[MASK]} tokens, and then the generator, a masked language model, fills \texttt{[MASK]} tokens with its prediction. The discriminator is trained to identify which tokens are generated.  The encoder of the discriminator is then used as the pre-trained model for fine-tuning on downstream tasks.
We summarize the extended token classification pre-training task in Figure~\ref{fig:framework}. 

\subsection{Extended Query Generation}

Each query is a sequence of tokens $\bx = [x_1, ..., x_n]$, where the number of the tokens $n$ is usually small for search queries. As a result, masked language models must set a high enough masking rate to make sure at least one token is masked out and the training can be really conducted with this sample. By replacing tokens with mask tokens, the semantic meaning of the search queries might be altered. 
Instead of masking tokens, we propose to insert \texttt{[MASK]} tokens in the query and use a generator to fill in the blanks. 
Specifically, we randomly select a set of positions $\bbm = [m_0, ..., m_n]$ with a fixed probability $p$: 
\begin{align*}
    m_i \sim {\rm Binomial}(p), ~ {\rm for} ~ i \in [0,...,n],
\end{align*}
where $m_i=1$ is the selected position. For each selected position $m_i=1$, we insert \texttt{[MASK]} between $x_i$ and $x_{i+1}$ \footnote{If $i=0$ or $i=n$, the inserted \texttt{[MASK]} is at the beginning or the ending of the sentence respectively. } We denote the extended input with \texttt{[MASK]} as 
\begin{align*}
    \bx^{\rm temp} = {\rm INSERT}(\bx, \bbm, \texttt{[MASK]})
\end{align*}
\vspace{-0.2in}

The generator $G$ is a pre-trained masked language model. Given the extended $\bx^{\rm temp}$, $G$ outputs a probability for generating a particular token $\tilde{x}_t$ for all $x_t^{\rm temp} = \texttt{[MASK]}$: 

\vspace{-0.2in}
\begin{align*}
    \hat{x}_t \sim p_G( x_t |\bx^{\rm temp})
\end{align*}
\vspace{-0.25in}

We denote the final extended input as 
\vspace{-0.05in}
\begin{align*}
    \bx^{\rm extend} = {\rm INSERT}(\bx, \bbm, \hat{\bx})
\end{align*}
\vspace{-0.3in}

\subsection{Training the Discriminator}

The training objective of the discriminator is identifying if a token is generated by the generator or not given the entire extended query. We denote the binary labels as $\by = [y_1, ..., y_{n'}]$, where
$y_t = \ind(x_t^{\rm temp}=\texttt{[MASK]})$ and $n'$ is the length of $\bx^{\rm extend}$. The training of $D$ is conducted via minimizing the following training loss:
\begin{align*}
   \min_{\theta_D} \cL(\bx, \theta_D) := \EE \left[ \sum_{t=1}^{n'} - y_t {\rm log}(D_{\theta}(\bx^{\rm extend}, t))  \right. \\
    \left. - (1-y_t) {\rm log}(1-D_{\theta}(\bx^{\rm extend}, t)) \right] , 
\end{align*}
where $\theta_D$ denotes the parameters of the discriminator. We remark that during the training of {\ours}, we only train the discriminator and keep the generator as unchanged for better efficiency. 
\section{Experiments}


We conducted experiments in an experimental system from E-commerce search domain. This paper considers the multilingual tasks and data, unless it is clearly stated. The pre-training corpus have 14 languages in total: En, De, Fr, Jp, It, Es, Zh, Pt, Nl, Tr, Cs, Pl, Ar, Sv. All the languages of the downstream tasks are included in the above 14 languages. The statistics of the data are presented in Table~\ref{tab:data}.

All implementations are based on \textit{transformers} \citep{wolf2019huggingface}. The max sequence length is set as $128$ for all of the following experiments.  We use an Amazon EC2 virtual machine with 8 NVIDIA A100 GPUs to conduct the experiments. 

\begin{table}[!htb]
    \centering
    \begin{tabular}{c|@{ }cccc}
        Dataset & Train & Dev & Test & Language \\
         \hline
        Query  & 1.0B & 1.1M & 1.1M & 14\\
        NER & 586.7K & 28.6K & 84.4K  & 13 \\
        Media & 52.3M & 6.5M & 6.5M  & 9\\
        Help & 29.8M & 3.7M & 3.7M & 9\\
        Adult & 10.8M & 1.3M & 1.3M & 9\\
        Spelling & 88.7M & 89.9K & 89.9K & 1$^*$ \\
    \end{tabular}
    \caption{Number of samples of datasets and number of languages. $^*$: Although the spelling correction task is only in English, we still use the multilingual pre-trained model for this task. }
    \vspace{-0.1in}
    \label{tab:data}
\end{table}

\subsection{Pre-training}

We use an E-commerce search query corpus, which consists of 1 Billion queries, for model pre-training. 
We first train a tokenizer on the query corpus using WordPiece \cite{wu2016google}. We use a vocabulary of 150K tokens. The tokenizer applies lower case, unicode, and accent normalization to the text.

We adopt the following transformer architecture for the encoder \citep{vaswani2017attention}: the number of layers is $12$, the hidden dimension is $386$ and the intermediate size of the feed-forward layer is $1536$. The total number of parameters of the encoder is $79$M. Note that, the adopted structure is the same MiniLM model\cite{wang2020minilm}. Such an architecture is even smaller than the usual transformer-base \citep{vaswani2017attention}. We adopt such a small architecture is mainly due to the real word application consideration, where there are latency constraints.  We use the RAdam Optimizer \citep{liu2019variance} with $beta=(0.9,0.999)$, a learning rate of $10^{-4}$, a weight decay of $0.01$ and a dropout ratio of $0.1$. The batch size is $64$ per GPU and the number of gradient accumulation steps is $2$. We adopt the cosine learning rate schedule \citep{loshchilov2016sgdr}.

We first train a masked language model with the above encoder from scratch. The number of training steps is $1$M. After MLM training, we take the MLM model as the generator, and use the MLM encoder to initialize the encoder of the discriminator. And then we conduct {\ours} training for $1$M step. The sampling ratio $p$ is $15\%$ in {\ours}. Note that during the training  of {\ours}, we fix the generator and do not apply dropout to the generator.

\begin{table*}[!htb]
    \centering
    \begin{tabular}{c|c|c|c|c|c}
       Model (\# of param.)  & NER: $F_1$ & Media: $F_1$ & Help: $F_1$ & Adult: $F_1$ & Spell Correction: Acc. \\
       \hline
       \multicolumn{6}{c}{Public Open-Domain Model}\\
       \hline
       MiniLM (117M) &  $69.45\%$ & $91.82\%$ & $88.88\%$ & $97.36\%$ & $76.00\%$ \\
       DistilBert (134M) & $68.73\%$ & $91.83\%$ & $88.07\%$ & $97.38\%$ & $76.81\%$  \\
       InfoXLM-Large (558M) & $73.29\%$ & $92.16\%$ & $89.52\%$ & $97.41\%$ & $77.00\%$  \\
       \hline
       \multicolumn{6}{c}{In-Domain Model}\\
       \hline
       MLM (79M) & $73.67\%$ & $92.51\%$ & $89.42\%$ & $97.55\%$ & $80.89\%$ \\
       ELECTRA (79M) & $72.26\%$ & $92.41\%$ & $89.08\%$ & $97.16\%$ & $80.61\%$ \\
       ETC (79M) & $\mathbf{74.23\%}$ & $\mathbf{92.62\%}$ & $\mathbf{89.95\%}$ & $\mathbf{97.71\%}$ & $\mathbf{81.17\%}$ \\
    \end{tabular}
    \caption{Main Experiment Results on $5$ QU tasks. All results obtained in this table are the average of 5 runs. We also did unpaired t-test between ETC and the second best method for all tasks. The improvement are statistically significant, where the p-value $<0.05$.}
    \vspace{-0.1in}
    \label{tab:main_result}
\end{table*}

\begin{table*}[!htb]
    \centering
    \begin{tabular}{c|c|c|c|c|c|c|c}
       Model  & Ar & Cs & De & En & Es & Fr & It \\
       \hline
       MLM & $72.76\%$ & $68.72\%$ & $79.07\%$ & $73.32\%$ & $73.75\%$ & $75.17\%$ & $73.16\%$  \\
       \hline
       ELECTRA & $72.95\%$ & $67.66\%$ & $78.32\%$ & $72.29\%$ & $72.73\%$ & $74.28\%$ & $72.34\%$ \\
       \hline
       ETC & $72.98\%$ & $68.94\%$ & $79.59\%$ & $73.82\%$ & $74.24\%$ & $75.69\%$ & $73.21\%$ \\
       ($F_1$ gain)& {\small $+0.22\%$} & {\small $+0.22\%$}  & {\small $+0.52\%$}  & {\small $+0.50\%$}  & {\small $+0.49\%$}  & {\small $+0.52\%$}  & {\small $+0.05\%$}  \\
       \hline
       \hline
       Model  & Jp & Nl & Pl & Pt & Sv & Tr & All\\
       \hline
       MLM & $73.26\%$ & $73.63\%$ &  $68.19\%$ & $76.97\%$ & $77.94\%$ & $78.37\%$ & $73.67\%$\\
       \hline
       ELECTRA & $71.75\%$ & $72.71\%$ & $66.99\%$ & $75.74\%$ & $75.59\%$ & $77.46\%$ & $72.26\%$ \\
       \hline
       ETC  & $74.94\%$ &  $73.95\%$ &  $68.84\%$ & $77.99\%$  & $79.51\%$ & $79.21\%$ & $74.23\%$\\
       ($F_1$ gain) & {\small $\mathbf{+1.68\%}$}  & {\small $+0.32\%$}  & {\small $+0.65\%$} & {\small $\mathbf{+1.02\%}$} & {\small $\mathbf{+1.57\%}$} & {\small $+0.84\%$}& {\small $+0.56\%$}
    \end{tabular}
    \caption{$F_1$ scores of different languages on NER task. For ETC, we also denote the performance gain over MLM in the second line. We highlight the languages where the performance gain is over $1\%$. }
    \label{tab:ner_multilang}
    \vspace{-0.1in}
\end{table*}

\subsection{Downstream Application}

To evaluate the quality of the pre-trained models, we fine-tune the pre-trained model in following five query understanding tasks:

\noindent~$\bullet$ \textit{Named Entity Recognition (NER)} is the task of detecting mentions of real-world entities from text and classifying them into predefined types (e.g., brand, color, product in E-commerce domain). There are $12$ different E-commerce entity types. It is a sequence labeling problem, we fine-tune the pre-trained encoder with a randomly initialized linear token-classification layer. We use the span-level $F_1$ score to evaluate the model performance. 

\noindent~$\bullet$ \textit{Media Query Identification} is the task of identifying weather the user is looking for a media product via the search query. As it is a binary unbalanced classification problem, we fine-tune the pre-trained encoder with a randomly initialized linear classification layer and use $F_1$ score to evaluate the model performance. 

\noindent~$\bullet$ \textit{Help Query Identification} is the task of identifying non-product related questions from other search queries. As it is a binary unbalanced classification problem, we fine-tune the pre-trained encoder with a randomly initialized linear classification layer and use $F_1$ score to evaluate the model performance. 

\noindent~$\bullet$ \textit{Adult Query Identification} is the task of identifying weather the user is looking for adult product. As it is a binary unbalanced classification problem, we fine-tune the pre-trained encoder with a randomly initialized linear classification layer and use $F_1$ score to evaluate the model performance. 

\noindent~$\bullet$ \textit{Spelling Error Correction} is the task of correcting the spelling errors in the search queries. It is a text generation task, we use a non-auto-regressive model which consists of the pre-trained encoder and a linear classification layer with the label space as the entire vocabulary. We use accuracy (the percentage of correctly fixed queries) as the metric to evaluate the model performance.

\textbf{Hyper-parameters}: the number of epoches is $10$ for the NER task and $2$ for other tasks. The batchsize is $128$ per GPU for the spelling error correction task, and $256$ per GPU for other tasks. We use the RAdam Optimizer \citep{liu2019variance}. The learning rate is selected from $\{2\times10^{-5}, 5\times10^{-5}, 5\times10^{-5}, 1\times10^{-4}, 2\times10^{-4}\}$ according to the validation set performance. We do not apply weight decay during fine-tuning. The dropout ratio is $0.1$.

\subsection{Baselines}

We compare {\ours} with the following open-domain multilingual pre-trained models: 

\noindent~$\bullet$ \textit{Multilingual DistilBERT} \citep{Sanh2019DistilBERTAD}

\noindent~$\bullet$ \textit{Multilingual MiniLM} \citep{wang2020minilm}

\noindent~$\bullet$ \textit{InfoXLM-Large} \citep{chi-etal-2021-infoxlm}

We also compare {\ours} with the following in-domain pre-trained models: 

\noindent~$\bullet$ \textit{MLM} is the  masked language modeling pre-training \citep{devlin2018bert} on the search query domain. 

\noindent~$\bullet$ \textit{ELECTRA} is the reproduction of \citet{clark2020electra} on the in domain query data. In our reproduction, we fix the generator, which is a pre-trained MLM model. The encoder of the discriminator is also initialized from the pre-trained MLM model. 

\noindent~$\bullet$ \textit{ETC} is our method. The same as the ELECTRA model, we take the MLM model as the generator, which is a pre-trained MLM model. The encoder of the discriminator is also initialized from the pre-trained MLM model. 

All in-domain pre-trained models are pre-trained from scratch and only use the query data. The tokenizer is also the same for all in-domain pre-trained models. For fair comparison the in-domain MLM model is trained for $2$M steps. Note that the MLM generator used in ELECTRA and ETC are only trained for $1$M steps, and the discriminator is trained for another $1$M steps.

\subsection{Main Results}

Our main results are shown in Table~\ref{tab:main_result}. 
First of all, the in-domain pre-trained MLM models outperform all the open-domain pre-trained models, including the InfoXLM-large, which has 550M parameter. Such a comparison indicates that our in-domain baselines are very strong. 

Among all the in-domain pre-trained models, {\ours} achieves the best performance in all 5 tasks. It's worthy noticing that ELECTRA perform even worse than the MLM model. 
\citet{clark2020electra} claims that ELECTRA can improve the pre-training mainly because two reasons: 1. the task is defined over all input tokens rather than just the small subset that was masked out, and 2. the binary classification task is easier than the language modeling task. 
As we show here, that is not the case of pre-training on short search queries. 
{\ours} and ELECTRA are similar in terms of both aspects: 1. applying loss on all tokens in the sequence, and 2. reducing the entire label space from the entire vocabulary to a binary classification. Unlike ELECTRA hurting the performance, {\ours} can improves the performance.
Such a comparison demonstrates that the benefit of {\ours} mainly comes from extending the short text.

\subsection{Training Efficiency}

\begin{figure}[!h]
    \centering
    \includegraphics[width=0.45\textwidth]{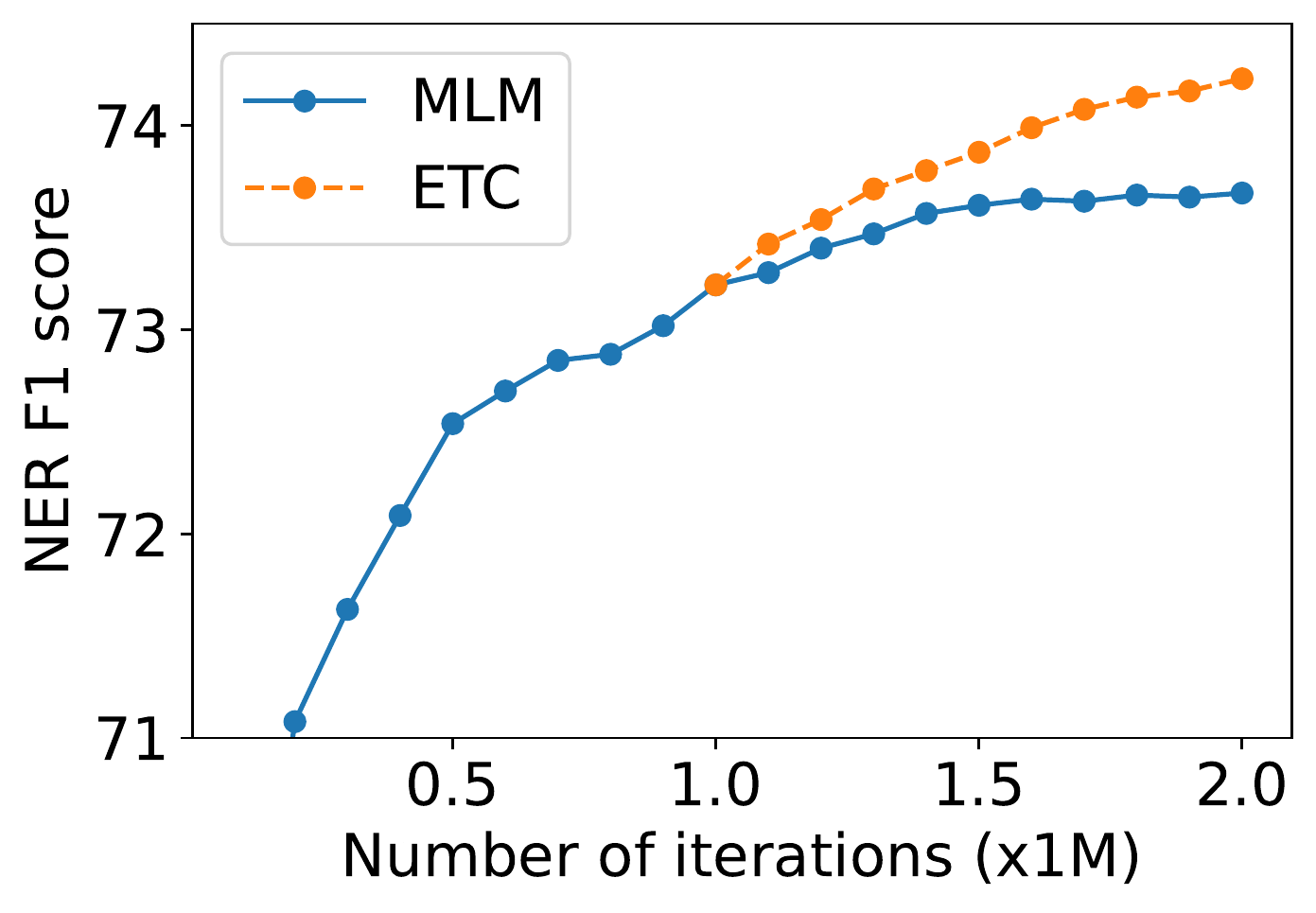}
    \vspace{-0.1in}
    \caption{The fine-tuning performance on NER with pre-trained models on difference pre-training iterations. }
    \label{fig:ner_perf}
    \vspace{-0.13in}
\end{figure}

We also study the training efficiency of the {\ours}. Specifically, we compare the fine-tuning performance on the NER task between MLM and {\ours} at different number of iterations. Note that MLM is trained for $2$M steps, while {\ours} is trained for $1$M and is continually trained from MLM checkpoint at $1$M step. 
The result is summarized in Figure~\ref{fig:ner_perf}. 

As can been seen, the NER $F_1$-score of MLM starts to saturated at around $1.5$M steps. Continual MLM can only give very limited performance gain. In contrast, {\ours} can keep improving the performance. 

\subsection{Multilingual Analysis}
\vspace{-0.05in}

We study how ETC would perform for different languages. Specifically, we exam the fine-tuning performance of different languages on the NER task, where there are $13$ difference languages. The results are presented in Table~\ref{tab:ner_multilang}.

As can be seen, ETC uniformly outperform MLM across all languages. ETC is particularly helpful for Jp, Pt, and Sv, where it achieves more than $1\%$ $F_1$ score improvement.

\subsection{Few-shot Experiments}

We conducted few-shot experiments on the NER task to demonstrate the effectiveness of {\ours}. Specifically, we fine-tune the pre-trained models on randomly sub-sampled training data and validate/test on the same validation/test set. The results are presented in Table~\ref{tab:few_shot_ner}. We observe that {\ours} outperforms the MLM pre-trained model on all splits of data. It achieves the best performance improvement on the $0.1\%$ data setting. Interestingly, we found the performance gains on $1\%$, $10\%$, $100\%$ are rather similar. Such a finding demonstrates that when scaling up the fine-tuning training data, the performance gain from {\ours} over MLM would not diminishes quickly. 

\begin{table}[!htb]
    \centering
    \begin{tabular}{@{}c@{ }|@{ }c@{ }|@{ }c@{ }|@{ }c@{ }|@{ }c@{}}
       Model  & $0.1\%$ & $1\%$ & $10\%$ &  $100\%$  \\
       \hline
       \# Samples & $0.6K$ & $5.9K$ & $58.6K$ & $586.7K$ \\
       \hline
       MLM  & $61.90\%$ & $66.36\%$ & $70.73\%$ & $73.67\%$ \\
       \hline
       {\ours}  & $62.74\%$ & $66.91\%$ & $71.23\%$ & $74.23\%$ \\
       ($F_1$ gain)  & {\small $+0.84\%$} & {\small $+0.55\%$} & {\small $+0.50\%$} & {\small $+0.56\%$} \\
       
    \end{tabular}
    \caption{$F_1$ scores on NER task with randomly sub-sampled training data. The first row is the sub-sampling ratios. }
    \label{tab:few_shot_ner}
    \vspace{-0.1in}
\end{table}
\vspace{-0.1in}
\section{Related Work}
\vspace{-0.05in}

\textbf{Self-Supervised Pre-training} There are two main streams of pre-training language models: 1. language modeling (LM), such as GPT-2 and GPT-3 \citep{radford2019language,brown2020language}; 2. masked language modeling (MLM), such as BERT. MLM usually perform better for language understanding tasks, while LM usually perform better in language generation tasks. 

ELECTRA \citep{clark2020electra} recently draws a lot of attention. \citet{clark2020electra} propose to replace the masked language modeling task by replaced token detection task. It improves not only sample-efficiency, but also fine-tuning performance. Recently, \citet{chi2021xlm} applied ELECTRA to train multilingual pre-trained model. \citet{He2021DeBERTaV3ID} combines ELECTRA with the DeBERTa architecture and achieved SOTA performance on GLUE benchmark \cite{wang2019glue}. In this paper, we show that ELECTRA is not effective for search queries. Given the success of ELECTRA in general, such a finding is exceptional. It suggests that the success in general domain is not always transferable to a specific domain.

\textbf{Query Understanding} QU is an important task since the appearance of search engine \citep{moore1995combining,lau1999patterns,boldasov2002user}. Some typical tasks are query intent classification, named entity recognition, ontology linking, spelling correction and query reformulation. 
The earliest systems heavily rely on domain lexicons and hand-crafted features (e.g., regular expression, grammar rules) \citep{dowding1994gemini}. Thanks to the advance in computing power, deep learning becomes trending in the QU \citep{nigam2019semantic,lin2020light}. More recently, pre-trained language model has been widely adopted in QU \citep{jiang2021named, kumar2019shareable, zhang2021queaco, li2021metats}. These work only adopts models pre-trained in open-domain, e.g., BERT. Some recent works have developed pre-training methods for E-commerce domain \citep{zhang2020bert,zhu2021knowledge}, but they are designed for understanding the long documents, e.g., product description, and none of them is particularly designed for query, which is short text.

The advances of QU is usually lagged behind of other NLP tasks since the lack of availability of related data in academic community. 
We hope this paper could inspire more research in this domain.

\vspace{-0.1in}
\section{Conclusion}
\vspace{-0.1in}

In this paper, we propose a new pre-training task -- Extended Token Classification ({\ours}) to learn representation for short text, such as search queries. 
Different from existing pre-training task, {\ours}  takes into consideration the short length of the text and improves the pre-training efficiency.
Our thorough experiments in E-commerce search domain demonstrate that
{\ours} outperforms existing methods in terms of fine-tuning performance on 5 QU tasks.

\section*{Ethical Impact}
{\ours} is a general framework for pre-training on short text, such as serach queries. 
{\ours} neither introduces any social/ethical bias to the model nor amplify any bias in the data. In all the experiments, we use internal data on an E-commerce search platform without knowing customers' identity. No customer/seller specific-data is disclosed. We build our algorithms using public code bases (transformers and PyTorch). We do not foresee any direct social consequences or ethical issues.

\bibliography{anthology,ref}
\bibliographystyle{acl_natbib}



\end{document}